\pdfoutput=1
\documentclass[11pt]{article}

\usepackage[]{emnlp2021}

\usepackage{times}
\usepackage{latexsym}

\usepackage[T1]{fontenc}
\usepackage[utf8]{inputenc}

\usepackage{microtype}

\usepackage{soul}
\usepackage{url}
\usepackage{graphicx}
\usepackage{amsmath}
\usepackage{amsthm}
\usepackage{booktabs}
\usepackage{tabularx}
\usepackage{multirow}
\usepackage{amssymb}
\usepackage{latexsym}
\usepackage{eucal}
\usepackage{bm}
\usepackage{amsfonts}
\usepackage{subfigure}
\usepackage{enumerate}
\usepackage{enumitem}
\usepackage{color}
\usepackage{titlesec}
\usepackage{flushend}
\usepackage{multicol}
\usepackage{float}
\usepackage{colortbl}

\usepackage[linesnumbered,ruled,vlined]{algorithm2e}%[ruled,vlined]{  
\usepackage{algpseudocode}

\definecolor{blue0}{RGB}{45,92,183}
\definecolor{gray0}{RGB}{100,100,100}
\definecolor{green0}{RGB}{112,173,71}
\definecolor{orange0}{RGB}{221,81,84}

\makeatletter
\g@addto@macro\normalsize{%
  \setlength{\abovedisplayskip}{2pt plus 2.0pt minus 1pt}%
  \setlength{\belowdisplayskip}{1pt plus 2.0pt minus 1pt}%
  \setlength{\abovedisplayshortskip}{1pt plus 2.0pt minus 1pt}%
  \setlength{\belowdisplayshortskip}{1pt plus 1pt minus 1pt}%
}
\title{GMH: A General Multi-hop Reasoning Model for KG Completion}

\author{
Yao Zhang$^1$ \  \ Hongru Liang$^2$ \ Adam Jatowt$^3$ \ Wenqiang Lei$^4$ \\
 \textbf{Xin Wei}$^5$ \ \textbf{ Ning Jiang}$^5$ \ \textbf{Zhenglu Yang}$^{1}$\thanks{~~Corresponding author.}\vspace{1mm}\\
  \normalsize{$^1$TKLNDST, CS, Nankai University, China, $^2$Sichuan University, China, } \\
  \normalsize{$^3$University of Innsbruck, Austria, $^4$National University of Singapore, Singapore, } \\
  \normalsize{$^5$Mashang Consumer Finance Co., Ltd., China} \\
  \normalsize{\texttt{yaozhang@mail.nankai.edu.cn, lianghongru@scu.edu.cn, }}\\ 
  \normalsize{\texttt{adam.jatowt@uibk.ac.at, wenqianglei@gmail.com, }} \\
  \normalsize{\texttt{$\left \{ \right .$xin.wei02, ning.jiang02$\left.\right \}$@msxf.com, yangzl@nankai.edu.cn}} \\}

\begin{document}
\maketitle
\begin{abstract}
Knowledge graphs are essential for numerous downstream natural language processing applications, but are typically incomplete with many facts missing. This results in research efforts on multi-hop reasoning task, which can be formulated as a search process and current models typically perform short distance reasoning. However, the long-distance reasoning is also vital with the ability to connect the superficially unrelated entities. To the best of our knowledge, there lacks a general framework that approaches multi-hop reasoning in mixed long-short distance reasoning scenarios.
    We argue that there are two key issues for a general multi-hop reasoning model: {\romannumeral1}) \textit{where to go}, and {\romannumeral2}) \textit{when to stop}. Therefore, we propose a general model which resolves the issues with three modules: 1) the local-global knowledge module to estimate the possible paths, 2) the differentiated action dropout module to explore a diverse set of paths, and 3) the adaptive stopping search module to avoid over searching.
    The comprehensive results on three datasets demonstrate the superiority of our model with significant improvements against baselines in both short and long distance reasoning scenarios. 
\end{abstract}

\section{Introduction}
\label{sec:introduction}

\begin{figure*}[t]
    \setlength{\belowrulesep}{-3pt}
    \setlength{\aboverulesep}{-3pt}
    \centering
    \includegraphics [width=0.99\textwidth]
    {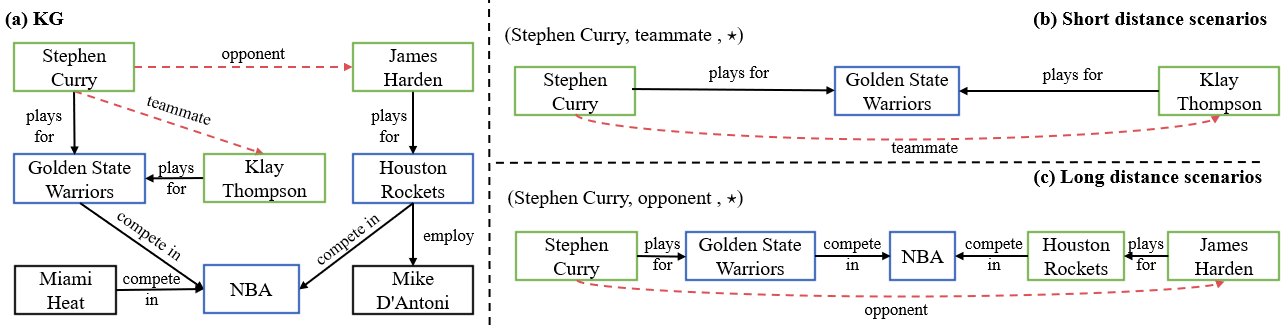}
    %\vspace{-2mm}
    \caption{Examples of (a) an incomplete knowledge graph, (b) a short distance scenario~(two-hop) about the reasoning of~\textit{(Stephen~Curry, teammate, $\star$)}, and (c) a long distance scenario~(four-hop) about the reasoning of~\textit{(Stephen~Curry, opponent, $\star$)}.
    The {\color{orange0}dotted} lines refer to the relations of incomplete triples and solid lines refer to existing relations.
    The {\color{green0}green}, {\color{blue0}blue} and 
    % {\color{gray0}gray} 
    black boxes represent the entities of the incomplete triples, the entities in the reasoning paths and the unrelated entities, respectively. As it can be seen, the long distance reasoning is needed and more complex than the short distance reasoning. Best viewed in color.}
    % \vspace{-2mm}
    \label{fig:intro_instance}      
    \end{figure*}

Knowledge graphs (KGs) have become the preferred technology for representing, sharing and adding factual knowledge to
% modern 
many natural language processing applications like recommendation~\cite{wang2019explainable,Lei2020Conversational}
% , conversational reasoning~\cite{moon-etal-2019-opendialkg} 
and question answering~\cite{xiao2019wsdm,zhang2018variational}.
% Automated reasoning, the ability of computing systems to create new inferences from observed evidence, has been a long standing goal of artificial intelligence~\cite{das2017go}.
% %{\color{blue}
% At present, various %of 
% natural language processing tasks perform
% %acquire factual knowledge through 
% automated reasoning in knowledge graphs~(KGs) to acquire factual knowledge, such as question answering~\cite{xiao2019wsdm} and conversational reasoning~\cite{moon-etal-2019-opendialkg}.
%}
KGs store triple facts~\textit{(head entity, relation, tail entity)} in the form of graphs, where entities are represented as nodes and relations are represented as labeled edges between entities~(e.g., Figure~\ref{fig:intro_instance} (a)).
% KG is a simple knowledge base which consists of facts in triple form~\textit{(head entity, relation, tail entity)}. 
% KG represents the triples in the form of graph, where entity is represented as the node and relation as the labeled edge between two entities.
Although popular KGs already contain millions of facts, e.g., YAGO~\cite{suchanek2007yago} and Freebase~\cite{freebase2008}, they are far from being complete considering the amount of existing facts and the scope of continuously appearing new knowledge. 
% generated due to rapid changes of the modern world.
% KG is never complete, even for a special domain, because its generation and updating methods often fail to handle adequate relevant facts.
% For example, as shown in Figure~\ref{fig:intro_instance}, the fact~~\textit{(Stephen~Curry, teammate, Klay Thompson)} is missing in the KG.
This has become the performance bottleneck of many KG-related applications, triggering
% which stimulated 
research efforts on the multi-hop reasoning task.
% The increasing concern with incompletion of KGs has stimulated research efforts on multi-hop reasoning task.
% The ability of finding some missing facts or reasoning on incomplete KGs is thus a worthy and challenge task.
% However, KGs are commonly highly incomplete because its generation and updating methods often fail to handle adequate relevant facts, %due to their lack of basic knowledge.
% which hampers the progress of automated reasoning. 
% %Therefore
% Accordingly, 

The multi-hop reasoning task can be formulated as a search process, in which the search agent traverses a logical multi-hop path to find the missing tail entity of an incomplete triple in KG.
% , given the head entity and relation.
% As shown in Fig.~\ref{fig:intro_instance}, to infer the incomplete triple~\textit{(Stephen~Curry, teammate,$\star$)}, where star denotes the missing tail entity, multi-hop reasoning finds the missing entity~\textit{Klay~Thompson} through a two-hop path~
% \textit{Stephen~Curry}
% $\xrightarrow[]{plays~for}$
% \textit{Golden~State~Warriors}
% $\xleftarrow[]{plays~for}$
% \textit{Klay~Thompson}.
As shown in Figure \ref{fig:intro_instance}
% (a) and 
(b), the two-hop path~
\textit{Stephen~Curry}
$\xrightarrow[]{plays~for}$
\textit{Golden~State~Warriors}
$\xleftarrow[]{plays~for}$
\textit{Klay~Thompson}
is searched to reason~\textit{Klay~Thompson} as the missing entity of
% in the incomplete triple
~\textit{(Stephen~Curry, teammate, $\star$)}, where $\star$ denotes the missing tail entity.
Multi-hop reasoning methods~\cite{Xiong2018DeepPath,das2017go} have been proposed to model the search process as a sequential decision problem in reinforcement learning~(RL) framework.
% In particular, MINERVA~\cite{das2017go} trains an model which uses the history path information to guide the search agent traversing in KG.
\cite{LinEMNLP2018} further optimized the reward function of RL framework based on~\cite{das2017go}.
% MultiHop~\cite{LinEMNLP2018} optimizes the reward function of RL framework based on MINERVA.
However, these works have only 
% investigated the multi-hop reasoning in the short distance scenarios
%我建议别在这写two three
scratched the surface of multi-hop reasoning as they focus only on short distance reasoning scenarios~(e.g., the two-hop case in Figure~\ref{fig:intro_instance} (b)).
% , i.e., two- or three-hop paths.

% Prior multi-hop reasoning models~\cite{Xiong2018DeepPath,das2017go} depend on the local path factor to tackle the first problem.
% The local path factor is far from sufficient to solve the problem of \textit{which path to select} because disposing a long distance case needs a global observation.
% And they all arbitrarily set the maximum number of search steps when dealing with the problem of \textit{when to stop search}.
% A fixed search step number is rigid to face the mixed short and long distance scenarios.

% Prior works has only scratched the surface of multi-hop reasoning because they only reasoning on two or three-hop paths.
% We observe that the \textbf{long distance reasoning scenarios} are the key point to promote the development of multi-hop reasoning and KG-related applications.
We observe that the \textbf{long distance reasoning scenarios} are vital in the development of multi-hop reasoning and KG-related applications, because two superficially unrelated entities may be actually deeply connected over a long distance.
With the significant expansion of KGs, the incompleteness of KG becomes more prominent, and long distance scenarios are rapidly increasing.
% Due to the significant expansion of KGs, long distance scenarios are rapidly proliferating. One should also keep in mind that two superficially unrelated entities may be actually deeply connected over a long distance. Restricting the search scope to a near area hampers the in-depth explorations of this rich underlying information, and thus, hurts the global reasoning performance. 
% Prior works showed satisfactory performance within 2 or 3 hops, nonetheless, they rarely analyzed the model capability of \textbf{long distance reasoning scenarios}\footnote{Considering the overall development of multi-hop reasoning, we denote the distances smaller or equal to 3 as ``short'' and the distances larger than 3 as ``long''.}.
As shown in Figure~\ref{fig:intro_instance} (c), the missing entity~\textit{James Harden} in the incomplete triple~\textit{(Stephen~Curry, opponent, $\star$)} is inferred by a long reasoning process, i.e., a four-hop path.
% Existing multi-hop reasoning models~\cite{Xiong2018DeepPath,chen2018variational,das2017go,LinEMNLP2018} show satisfactory performance %results in simple reasoning cases, i.e., only 
% within 2 or 3 hops, whereas long distance reasoning\footnote{Considering the overall development of multi-hop reasoning, we denote the distances smaller or equal to 3 as ``short'' and the distances larger than 3 as ``long''.} is neglected. 
% For another incomplete KG triple~$(Stephen~Curry, opponent, ?)$ in Fig.~\ref{fig:intro_instance},
% %As shown in Fig.~\ref{fig:intro_instance}, what is the relation between ``Stephen Curry" and ``James Harden"? It may be opponent or teammate. 
% the answer ``James Harden'' is inferred by a long reasoning process, that is, a four-hop path.
% With the aggressive expansion of KGs, long distance scenarios are rapidly proliferating, and two superficially unrelated entities may be deeply connected over a long distance. Restricting the search scope within a near area hampers in-depth explorations of this rich underlying information, and accordingly, compromises the global reasoning performance. As the proverb says :``throw a long line to catch a big fish.'' In this study, we expect to lengthen our search line and loosen the current restriction of reasoning hops, specifically toward long distance-aware multi-hop reasoning that has been scarcely touched before. 
Moreover, in practice, the long and short distance reasoning scenarios are mixed.
The ideal multi-hop reasoning model should be competent on mixed short and long distances.
Specifically, we argue that there are two key issues in the traverse of KG that need to be resolved:

% We argue that there are two key issues in the traverse of KG for 
% We observe that the multi-hop reasoning task in long distance scenarios is more challenging than that in the short distance scenarios because of two key issues in the traverse of KG:
% % because an effective long distance-aware multi-hop reasoning strategy should tackle the following two key problems when its search agent traverses in KG
% % :
\begin{itemize}[leftmargin=-1pt,topsep = -1 pt,partopsep= 1 pt,labelwidth=3pt,labelsep= 1 pt]
    \setlength{\itemsep}{0pt}
    \setlength{\parsep}{0pt}
    \setlength{\parskip}{0pt}
    % \item[i)] \textit{Which edge to select?} The search agent needs to select the positive edge to find the target~(missing) entity and to avoid selecting the negative edge which will cause the search agent to move away from the target entity. 
    % When the search distance increases, the selections faced by the search agent also increase exponentially; and the impact of selecting a negative edge will affect subsequent selections. Accordingly, the edge selection is more difficult in long distance reasoning scenarios.
    \item[{\romannumeral1})] \textit{Where to go?} The search agent needs to decide where to go at next search step, i.e., selecting an edge connected with the current node. 
    Selecting the positive edge means that the agent will move towards the target node, otherwise, it will move away from the target.
        When the search distance increases, the issue becomes more challenging because the agent needs to make more decisions.
   
    % The search agent needs to select the positive edge to find the target~(missing) entity and to avoid selecting the negative edge which will cause the search agent to move away from the target entity. 
    % When the search distance increases, the selections faced by the search agent also increase exponentially; and the impact of selecting a negative edge will affect subsequent selections. Accordingly, the edge selection is more difficult in long distance reasoning scenarios.
    \item[{\romannumeral2})] \textit{When to stop?} The search agent needs to decide when to stop the search because the exact search steps cannot be known in advance. An ideal search agent needs to stop at a suitable time to avoid over searching and adapt to realistic reasoning scenarios with mixed short and long distances.
\end{itemize}

\begin{figure*}[t]
    \setlength{\belowrulesep}{-3pt}
    \setlength{\aboverulesep}{-3pt}
\centering
\includegraphics [width=0.99\textwidth]
{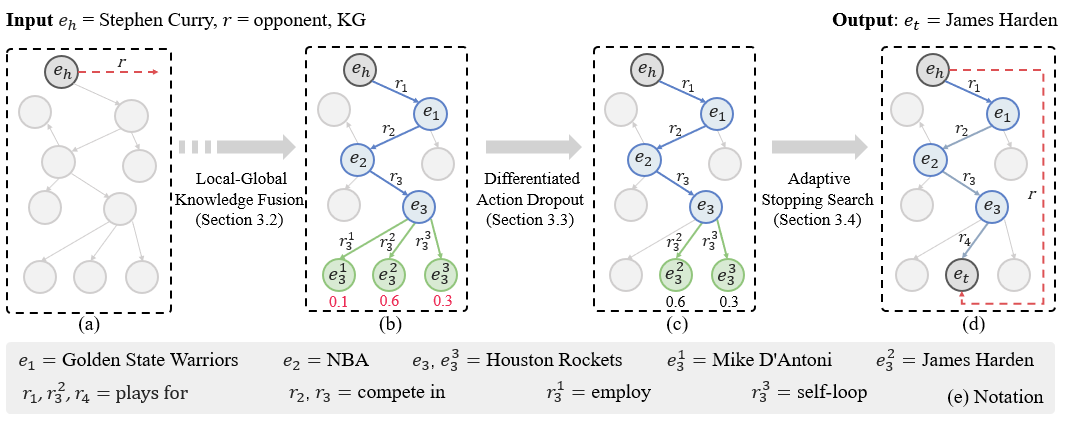}
% \vspace{-2mm}
\caption{An illustration of the GMH Model. 
We reuse the example in Figure\ref{fig:intro_instance} (c) for explanation.
The input includes the head entity and the relation of the incomplete triple \textit{(Stephen~Curry, opponent, $\star$)} with the background KG, and the output is the tail entity \textit{James Harden}. The subgraph (a) is the initial state of the search process. The subgraphs (b-d) show the search process at step 4. Specifically, 1) we develop the local-global knowledge fusion module to estimate the possible paths, 2) the differentiated action dropout module to dilute the negative paths, and 3) the adaptive stopping search module to avoid over searching. Best viewed in color.}
% \vspace{-3mm}
\label{fig:modeloverview}      
\end{figure*}
To this end,
we propose a \textbf{G}eneral \textbf{M}ulti-\textbf{H}op reasoning model, termed GMH, which solves the two above-listed issues in three steps:
1) the local-global knowledge fusion module fuses the local knowledge learnt from history path and the global knowledge learnt from graph structure;
% local-global factors fusion, which fuses the local path and global embedding factors to facilitate the agent to select the positive path;
2) the differentiated action dropout module forces the search agent to explore a diverse set of paths from a global perspective; 
and 3) the adaptive stopping search module uses a self-loop controller to avoid over searching and resource wasting.
We train the policy network with RL and optimize the reward to find the target entity effectively.
In summary, the main contributions of this work are as follows:
\begin{itemize}[leftmargin=-1pt,topsep = -1 pt,partopsep= 1 pt,labelwidth=3pt,labelsep= 1 pt]
    \setlength{\itemsep}{0pt}
    \setlength{\parsep}{0pt}
    \setlength{\parskip}{0pt}
    \item We observe that the long distance reasoning scenarios are vital in the development of multi-hop reasoning, and argue that an ideal multi-hop reasoning model should be competent on mixed long-short distance reasoning scenarios.
    \item We propose a general multi-hop reasoning model, GMH, which can solve two key issues in mixed long-short distance reasoning scenarios: i) where to go and ii) when to stop.

    % \item We design the local-global knowledge module to integrate knowledge of history paths and graph structure, the differentiated action dropout module to diversify the reasoning paths, and the adaptive stopping search module to define the optimal steps of searching.

    \item We evaluate GMH on three benchmarks, FC17, UMLS and WN18RR. The results demonstrate the superiority of GMH with significant improvements over baselines in mixed long-short distance reasoning scenarios and with competitive performances in short distance reasoning scenarios. 
\end{itemize}

\section{Related Work}
\label{sec:related work}

% \clearpage
In this section, we summarize the related work and discuss their connections to our model.
Firstly, we introduce the two lines of work on the KG completion task: multi-hop reasoning and KG embedding.

The \textbf{multi-hop reasoning} task focuses on learning logical multi-hop paths reasoned from KG.
% Multi-hop reasoning task can be formulated as a search process, in which the search agent traverses a logical multi-hop path to find the missing tail entity in KG, based on the known head entity and relation.
The multi-hop reasoning models distill deep information from paths thereby generating further directly interpretable results.
\cite{lao2011random,das2016chains,jiang2017attentive,yin2018recurrent}
% ,~\cite{}
% ,~\cite{}, and~\cite{} 
predicted the missing relations of incomplete triples based on pre-computed paths. 
\cite{Xiong2018DeepPath} firstly adopted the RL framework to improve the reasoning performance. 
% finds paths between the head and tail entities based on RL framework and predicts the missing relations.
% Our work is devised to find the missing entities, which are orthogonal to the aforementioned models predicting missing relation in a complementary manner.
% As for finding a missing entity, 
The task of finding a missing entity is orthogonal to the prediction of the missing relation in a complementary manner.
\cite{das2017go} used the history path to facilitate the search agent finding the missing entity
% effectively 
and \cite{LinEMNLP2018} optimized the reward function of RL framework based on \cite{das2017go}.
% MINERVA~\cite{das2017go} uses the history path to facilitate the search agent to find the missing entity effectively.
% MultiHop~\cite{LinEMNLP2018} optimizes the reward function of RL framework based on MINERVA.
\cite{lv-etal-2019-adapting} adopted the meta learning framework for multi-hop reasoning over few-shot relations.
% However, these works neglect the long distance reasoning scenarios, their performance declines significantly as the distance increases.
These works are conditioned in short distance scenarios, and tend to rapidly lose effectiveness as the distance increases. In contrast, we propose a general model which can be sufficiently utilized in both the short and long distance reasoning scenarios.

The \textbf{KG embedding} task is another line of work carried to alleviate the incompleteness of KG.
Embedding-based models project KGs in the embedding space and estimate the likelihood of each triple using scoring functions.
% , through learning a scoring function to indicate the predicted probability.
% learn global latent representations for entities and relations in continuous
% vector space and estimate the likelihood of each triple (e s ,r,e d ) by using a scoring function ψ(e s ,r,e d ).
% TransE~\cite{bordes2013translating}, based on translation assumption, considers that two entities can be connected by relation with low error. TransH~\cite{wang2014knowledge}, TransR~\cite{lin2015learning}, and TransD~\cite{ji2016knowledge} are the variants of TransE. 
\cite{bordes2013translating,wang2014knowledge,lin2015learning,ji2016knowledge} defined additive scoring functions based on the translation assumption.
% namely, two entities can be connected by the relation with the lowest error.
\cite{yang2014embedding,trouillon2017complex} defined multiplicative scoring functions based on linear map assumption.
Moreover, recent models introduce special neural networks like neural tensor network~\cite{NIPS2013_tensor}, convolution neural network~\cite{Dettmers2018Convolutional} and graph convolutional network~\cite{nathani-etal-2019-learning}.
% For example, neural tensor networks in~\cite{NIPS2013_tensor}, convolution neural network in~\cite{Dettmers2018Convolutional} and graph convolutional network in~\cite{nathani-etal-2019-learning}.
Due to the neglection of deep information within multi-hop paths, the results of the embedding-based models lack interpretability, which is critical for KG-related applications. 
However, embedding-based models are less sensitive to the reasoning distance because they learn KG structure from the global perspective.
Thus, we take advantage of this strength to learn the global knowledge from graph structure and retain the interpretability by reasoning from the history paths. 
% However, it is critical to understand how models behave when exposed to real-world data, especially for KG-related applications.

%\vspace{-1mm}

Secondly, we discuss the community research on long distance reasoning scenarios.
% Nevertheless, t
% There are studies that focus on long distance reasoning scenarios in other natural language processing applications.
% For example, in conversational reasoning task,
\cite{tuan-etal-2019-dykgchat} formed a transition matrix for reasoning over six-hop path in KG for the conversational reasoning task.
It is however not suitable for large-scale KGs, because the matrix multiplication requires large calculation space.
\cite{wang2019explainable} proposed a long-term sequential pattern to encode long distance paths
% and estimate the recommendation confidence 
for the recommendation task.
Because there is no real reasoning process for the long distance paths, it is not suitable for the KG completion.
To summary, we are the first to study long distance reasoning scenarios in the KG completion.
We propose a general model that tackles both short and long distance reasoning scenarios and works effectively on large-scale KGs.

%\clearpage

\section{Methodology}
\label{sec:model}

Figure~\ref{fig:modeloverview} illustrates the entire process of the GMH model. The input involves the head entity and relation of the incomplete triple with the background KG. The output is the missing tail entity.
We systematize the model in three steps:~1)~the local-global knowledge fusion module to integrate knowledge of history paths and graph structure;~2)~the differentiated action dropout module to diversify the reasoning paths; and~3)~the adaptive stopping search module to formulate the optimal steps of searching.
% The local-global factors fusion module fuses the local path and global embedding factors to facilitate the agent to select the positive path.
% The differentiated path dropout module forces the agent to explore a diverse set of paths from a global perspective.
% The adaptive stopping search uses a self-loop controller to avoid over searching and resource wasting.
The local-global knowledge fusion and differentiated action dropout modules facilitate the agent to address the issue of \textit{where to go}. The adaptive stopping search module controls the search steps to resolve the issue of \textit{when to stop}.

\subsection{Preliminary}
% \subsubsection{Definition}

% Before going to the detailed model description, 
We formally represent a KG as a collection of triples~$\mathcal{T}=\left \{ (e_{h},r,e_{t})|e_{h}\in \mathcal{E}, e_{t}\in \mathcal{E},r\in \mathcal{R}\right \} $, where $e_{h}$, $r$ and $e_{t}$ denote the head entity, relation and tail entity in one triple, $\mathcal{E}$ and $\mathcal{R}$ are the entity and relation sets, respectively. 
Each directed link in KG represents a valid triple~(i.e., $e_{h}$ and $e_{t}$ are represented as the nodes and $r$ as the labeled edge between them).
For an incomplete triple, multi-hop reasoning can be perceived as searching a target tail entity~$e_{t}$ through limited steps in KG, starting from head entity $e_{h}$ and based on the relation $r\in \mathcal{R}$. 
We use query~$q$ to represent $(e_{h},r)$ in the following sections.
At step $s$, the search agent will transfer to the entity~$e_{s}$ updating 
the history path trajectory~$\mathcal{H}_{s}=\left \{e_{h}, r_{1}, e_{1},..., r_{s}, e_{s}\right \}$, and the available action set  $\mathcal{A}_{s}=\left \{ ({{r_{s}^{i}}},{{e_{s}^{i}}} )|({e}_{s},{{r_{s}^{i}}},{{e_{s}^{i}}} )\in \mathcal{T}\right \}$.
%from which the agent will select one in the subsequent step. 
$\mathcal{A}_{s}$ consists of all outgoing relations and the associated entities of~${e}_{s}$.
The agent will select one action from~$\mathcal{A}_{s}$ to transfer to the next entity~${e}_{s+1}$ through the correlated relation~${r}_{s+1}$ at next step.

\subsection{Local-Global Knowledge Fusion}

\begin{figure}[t]
        \setlength{\abovecaptionskip}{0.cm}
                \setlength{\belowcaptionskip}{-0.1cm}
\centering
\includegraphics [width=0.39\textwidth]
{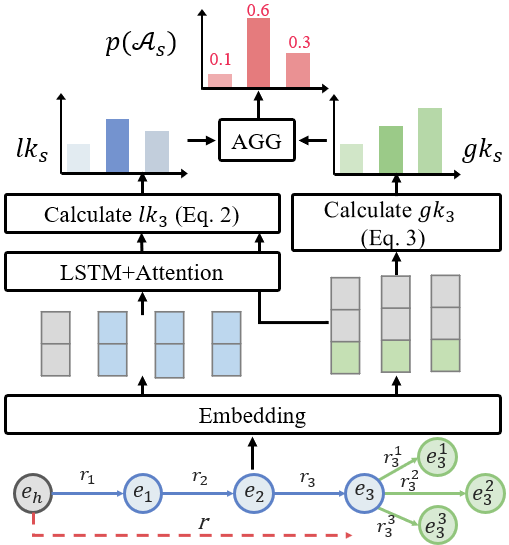}
\caption{An illustration of the local-global knowledge fusion module. We reuse the search process in Figure 2 for detailed explanation. Best viewed in color.
% The left part learns the local knowledge
% from 
% % through encoding 
% the history path trajectory. 
% The right part learns the global knowledge from graph structure through a pre-trained embedding-base model. The fusion is conducted in the middle part and produces the probability distribution of
% % the final score for 
% the available actions.
% % Best viewed in color.
}
%\vspace{{-2mm}
\label{fig:fusion}      
\end{figure}

% \noindent
% \textbf{Fusing Local Path and global knowledges}
% \label{sec:Combine_his_tem}
In this module, we learn local knowledge~$lk_{s}$ and global knowledge~$gk_{s}$ to resolve the ``where to go" issue,
% assist the agent in effectively finding the target tail entity, 
as shown in Figure~\ref{fig:fusion}.
The local knowledge indicates that the agent makes decisions on the basis of the history path trajectory~$\mathcal{H}_{s}$ at step $s$ from a local perspective. 
The global knowledge is calculated through a pre-trained embedding-based models from a global perspective.
We use an aggregate~(abbr. AGG) block to aggregate $lk_{s}$ and $gk_{s}$, which has two types: summation~($lk_{s}+gk_{s}$) and scalar product~($lk_{s}*gk_{s}$).
The distribution $p(\mathcal{A}_{s})\in \mathbb{R}^{\left | \mathcal{A}_{s} \right |}$ is calculated through the AGG block and represents the confidence score for each available entity in $\mathcal{A}_{s}$.
% , and it lies on simplex of size $\left | \mathcal{A}_{s} \right |$.
The agent will select one action from~$\mathcal{A}_{s}$ according to the distribution $p(\mathcal{A}_{s})$
to transfer to the next entity.

\noindent
\textbf{Local Knowledge Learning}
% \label{sec:HTF}

The local knowledge~$lk_{s}$ indicates  from a local perspective that the agent makes decisions based on the history path trajectory~$\mathcal{H}_{s}$ at step $s$. We adopt long short-term memory~(LSTM) neural network and attention mechanism  to encode the history path trajectory and yield the local knowledge. 
% We expect that $lk_{s}$ contributes to improving the accuracy of the agent's decisions. 

The history path trajectory~$\mathcal{H}_{s}=(e_{h}, r_{1}, e_{1},..., r_{s}, e_{s})$ consists of the sequence of entities and relations which the agent has selected over the last $s$ steps. 
%Every entity and relation in $\mathcal{G}$ are assigned dense vector embeddings ${v_{e}}\in \mathbb{R}^{dim}$ and ${v_{r}}\in \mathbb{R}^{dim}$ by a look-up operation, where $dim$ is the dimension.
We adopt an embedding layer to generate the embedding of entities and relations.
The embedding of query is~$\vec{q}=[\vec{e_{h}};\vec{r}] \in \mathbb{R}^{2dim}$, i.e., the concatenation of the embeddings of the head entity $\vec{e_{h}}\in \mathbb{R}^{dim}$ and relation $\vec{r}\in \mathbb{R}^{dim}$, where $dim$ is the dimension.
We use an LSTM to encode the embedding of $\mathcal{H}_{s}$ to yield the hidden state embedding sequence $(\vec{h}_{0},...,\vec{h}_{s})$, where~$\vec{h}_{s}=LSTM(\vec{h}_{s-1},[{\vec{r_{s}}},{\vec{e_{s}}}])\in \mathbb{R}^{2dim}$ is the hidden state at step $s$,  ${e_{s}}$ is the current entity and ${r_{s}}$ is the relation that connects ${e_{s-1}}$ and ${e_{s}}$.

Prior works~\cite{das2017go,LinEMNLP2018} use only the current hidden state embedding~(i.e., $\vec{h}_{s}$) to yield the next action and they neglect the differentiated importance between the hidden states over the last $s$ steps.
Therefore, the attention weight value calculated between the hidden state embedding sequence and the query embedding is introduced to optimize the local knowledge~$lk_{s}$.
Each weight value is derived by comparing the query ${\vec{q}}$ with each hidden state~$\vec{h}_{i}$: 
\begin{equation}
\alpha({\vec{q}},\vec{h}_{i}) =\frac{\exp (f({\vec{q}},{\vec{h}_{i}}))}{\sum_{j=0}^{s}\exp (f({\vec{q}},{\vec{h}_{j}}))},
\end{equation}
where $i$ and $j$ stand for the $i$-th and $j$-th hidden state candidate, respectively.
Here, $f(\cdot )$ is represented as a query-based function: $f({v_{q}},{h_{m}})={v_{q}}^{\top }{h_{m}}$.
% \begin{equation}
% score({v_{q}},{h_{t}})={v_{q}}^{\top }{h_{t}}.
% \end{equation}
%The weight value reflects the internal correlation between the query $v_{q}$ and the history path trajectory $\mathcal{H}_{s}$. 
%The local knowledge~$lk_{s}$ reflects the influence of the history path trajectory on each element in the available action $\mathcal{A}_{s}$. 
Ultimately, local knowledge~$lk_{s} \in \mathbb{R}^{\left | \mathcal{A}_{s} \right |} $, which reflects the influence of the history path trajectory on each element in $\mathcal{A}_{s}$, can be obtained:
\begin{equation}
%HTF_{t}  =\sum_{t\in T}\alpha _{t}(r_{q})h_{t},
\vec{lk}_{s}  ={\mathcal{A}}_{s}\times {\bm{W}}_{1}\delta _{1}({\bm{W}}_{2}\sum_{m=1}^{s}\alpha({\vec{q}},\vec{h}_{m})\vec{h}_{m}),
\label{eq:lpf}
\end{equation}
\noindent where 
%${\mathcal{A}}_{t}$ is the available action set at step $s$, 
${\bm{W}}_{1}$ and ${\bm{W}}_{2}$ are the weights, and $\delta _{1}$ is the activation function.

\noindent
%\vspace{{-2mm}
\textbf{Global Knowledge Learning}
\label{sec:GEF}

Prior works~\cite{das2017go,LinEMNLP2018} use the local knowledge and neglect the long distance cases which requires higher decision accuracy of the agent.
We introduce the global knowledge~$gk_{s}$ learnt from graph structure by a pre-trained embedding-based model.
% adopt the existing embedding-based models to calculate the global knowledge~$gk_{s}$ from graph structure.

%The global knowledge~$gk_{s}$ indicates that the agent makes decisions on the basis of the query~$(e_{h},r_{q})$ from a global perspective by using existing embedding-based models.

Embedding-based models map the graph structure in continuous vector space by using a scoring function~$\psi(e_{h},r,e_{t})$.
% to maximize the score of each triple~$(e_{h},r,e_{t})$.
% learn global latent representations for entities and relations in continuous vector space and estimate the likelihood of each triple~$(e_{h},r,e_{t})$ by using a scoring function~$\psi(e_{h},r,e_{t})$.
We generate the new triple~$(e_{h},r,e_{s}^i)$ by concatenating the head entity and relation with available entity $e_{s}^i \in {\mathcal{E}}_{t}^{\mathcal{A}}$, where ${\mathcal{E}}_{t}^{\mathcal{A}}\in \mathbb{R}^{\left | \mathcal{A}_{s} \right |\times dim}$ contains all available entities in $\mathcal{A}_{s}$.
As we consider that the positive available entity is closer to the target tail entity in vector space, combining the positive available entity in $\mathcal{A}_{s}$ with the query will get a higher score than that using negative available entities.
% Therefore, we generate the new triple~$(e_{h},r,e_{s}^i)$ by concatenating the head entity and relation with $e_{s}^i \in {\mathcal{E}}_{t}^{\mathcal{A}}$, where ${\mathcal{E}}_{t}^{\mathcal{A}}\in \mathbb{R}^{\left | \mathcal{A}_{s} \right |\times dim}$ contains all available entities in $\mathcal{A}_{s}$.
% We adopt an embedding layer to generate the embedding of each new triple.
Formally, we adopt a pre-trained embedding-based model
% scoring function~$\psi(\cdot)$ 
to calculate these new triples to obtain the global knowledge~$gk_{s}$:
%The difference with the local knowledge~$HTF_{t}$ is that the global knowledge~$TTF_{t}$ relies on the similarity between the available action with the query entity and the query relation in the continuous vector space.
%Formally, the global knowledge~$TTF_{t}$ is defined as follows:
\begin{equation}
        \vec{gk}_{s} =[\psi({\vec{e}_{h}},\vec{r},{\vec{e}_{s}}^{~1});...;\psi({\vec{e}_{h}},\vec{r},\vec{e}_{s}^{~\left | \mathcal{A}_{s} \right |})].
\label{eq:gef}
\end{equation}
% where $N_{A}$ is the number of the available actions. 

Concatenating each of new triples' scores gives the global knowledge~$gk_{s} \in \mathbb{R}^{\left | \mathcal{A}_{s} \right |}$. The selection of scoring function~$\psi(\cdot)$ is discussed in Section~\ref{sec:ablation}.

\subsection{Differentiated Action Dropout}
\label{sec:actiondropout}
In the multi-hop reasoning task, it is important to enforce effective exploration of a diverse set of paths and dilute the impact of negative paths.
\cite{LinEMNLP2018} forced the agent to explore a diverse set of paths using action dropout technique which randomly masks some available actions in $\mathcal{A}_{s}$, i.e., blocking some outgoing edges of the agent. 
% Yet, that approach cannot discriminate paths of different qualities.
However, in the case of reasoning over long distances, the number of paths is much greater than that in the short distance scenarios because the search space grows exponentially.
The random action dropout technique is inefficient because it cannot discriminate paths of different qualities.
We then propose the differentiated action dropout~(DAD) technique based on the global knowledge~$gk_{s}$ to mask available actions, 
%Because we consider 
since we believe that higher-scoring actions are more likely to exist in a high-quality path.
In particular,
%In contrast, we use the global embedding factor~$GEF_{t}$ as the probability which is related to per available action. 
the mask matrix~$M_t \in \mathbb{R}^{\left | \mathcal{A}_{s} \right |}$ is sampled from the Bernoulli distribution:
\begin{equation}
        \vec{M}_t \sim  Bernoulli(sigmoid(\vec{gk}_{s})).
\label{eq:dropout}
\end{equation}

The element in $M_t$ is binary, where 1 indicates the action is reserved and 0 indicates abandonment.
The fusion of local-global knowledge and differentiated action dropout modules helps the agent to tackle the key problem~\textit{where to go} jointly.

%\vspace{{-2mm}
\subsection{Adaptive Stopping Search}
\label{sec:earlystopping}
For the second key issue of~\textit{when to stop}, 
we devise the adaptive stopping search~(ASS) module inspired by the early stopping strategy~\cite{Prechelt97earlystopping} which is used to avoid overfitting when training a learner with an iterative method.
We add a self-loop action~$($\textit{self-loop}$,e_{s})$ to give the agent an option of not expanding from~$e_{s}$.
When the agent chooses the self-loop action for several times, we consider it means that the agent has found the target tail entity, thus it can choose to end early.

In this module, we devise a self-loop controller to avoid over searching and resource wasting.
The self-loop controller has a dual judgment mechanism based on the the maximum search step~$S$ and the maximum loop number~$N$.
When the search step reaches the maximum~$S$, or the agent selects the self-loop action for $N$ consecutive times, the search process will be stopped.
Using the ASS strategy improves our model's scalability on both short and long distances and effectively avoids wasting of resources caused by over-searching.

\begin{algorithm}[t]
        \footnotesize
        \caption{Training process of GMH}
        \label{algorithm:training}
        \LinesNumbered
        \KwIn{The training samples set~$\mathcal{D}_{train}$;
                        background KG~($\mathcal{T}$, $\mathcal{E}$, $\mathcal{R}$);
                        the maximum search step~$S$;
                        the maximum loop number~$N$;
                        the randomly initialized parameters $\theta$
                        }
      \KwOut{The optimized parameters~$\theta$}

      \Repeat  {model converged}
      {
        Sample a triple $(e_{h},r,e_{t})$ from $\mathcal{D}_{train}$;
      
      \textbf{Initialize}: $s=0$; $n=0$;
      $e_{s}=e_{h}$; $\mathcal{H}_{s}=\left \{e_{h}\right \}$; $\mathcal{A}_{s}=\left \{ ({{r_{s}^{i}}},{{e_{s}^{i}}} )|({e}_{h},{{r_{s}^{i}}},{{e_{s}^{i}}} )\in \mathcal{T}\right \}$; 
      
      \For {$s<S$ and {$n<N$}}
      {
        % Calculate the local and global knowledge~$lk_{s}$~(Eq.~\ref{eq:lpf}) and the global knowledge~$gk_{s}$~(Eq.~\ref{eq:gef});
        Calculate $lk_{s}$~(Eq.~\ref{eq:lpf}) and $gk_{s}$~(Eq.~\ref{eq:gef});

        %        \State Calculate the global knowledge~$gk_{s}$~(Eq.~\ref{eq:gef})
        Fuse $lk_{s}$ and $gk_{s}$ to yield the final score;

        Dropout actions from~$\mathcal{A}_{s}$~(Eq.~\ref{eq:dropout});
        %  based on the mask matrix~$M_s$

        Select the next entity~$e_{s+1}$ and the related relation~$r_{s+1}$;
        
        \If {$r_{s+1}=$self-loop}{             $n\leftarrow n+1$;
}
                
        \textbf{update} $His_{s+1}\leftarrow \mathcal{H}_{s}\cup \left \{r_{s+1},e_{s+1}\right \}$; $\mathcal{A}_{s+1}\leftarrow \left \{ ({{r_{s+1}^{i}}},{{e_{s+1}^{i}}} )|({e}_{s+1},{{r_{s+1}^{i}}},{{e_{s+1}^{i}}} )\in \mathcal{T}\right \}$;
        
        $s\leftarrow s+1$;
      }
      $\hat{e_{t}}=e_{s}$; calculate the reward~$R(\hat{e_{t}}|e_{h},r,e_{t})$ and update $\theta$~(Eq.~\ref{eq:gradient});
      }
\end{algorithm}

%\vspace{{-2mm}
\subsection{Training}
\label{sec:training}

Following~\cite{das2017go}, we frame the search process as a Markov Decision Process~(MDP) on the KG
%: on the basis of the query relation~$r_{q}$, the initial state is $e_{s}$ is updated to a new entity by the agent's decision at each step until the state arrives at the target entity~$e_{d}$.
and adopt the on-policy RL method to train the agent. 
We design a randomized history-dependent policy network $\pi=(p(\mathcal{A}_{1}),...,p(\mathcal{A}_{s}),...,p(\mathcal{A}_{S}))$.
The policy network is trained by maximizing the expected reward over all training samples~$\mathcal{D}_{train}$:
\begin{equation}
\begin{aligned}
J(\theta)=&\mathbb{E}_{(e_{h},r,e_{t})\sim \mathcal{D}_{train}}\\
&[\mathbb{E}_{A_{1},...,A_{S}\sim \pi}[R(\hat{e_{t}}|e_{h},r,e_{t})]].
\end{aligned}
%%%%%\vspace{{-2mm}
\end{equation}

where $\theta$ denotes the set of parameters in GMH, $R(\cdot )$ is the reward function and $\hat{e_{t}}$ is the final entity chosen by the agent. If $\hat{e_{t}}={e}_{t}$, then the terminal reward is assigned +1 and 0 otherwise.

The optimization is conducted using the REINFORCE algorithm~\cite{reinforce1994} which iterates through all $(e_{h},r,e_{t})$ triples in $\mathcal{D}_{train}$ and updates $\theta $ with the following stochastic gradient:%%%%%\vspace{{-3mm}
\begin{equation}
        \bigtriangledown _{\theta }J(\theta )\approx \bigtriangledown _{\theta }\sum\nolimits_{s}R(\hat{e_{t}}|e_{h},r,e_{t})\log \pi _{\theta }.
\label{eq:gradient}    
\end{equation}

The training process is detailed in Algorithm 1. 
% It takes as input the training samples set, background KG, the maximum search step, the maximum loop number and the randomly initialized parameters.
During a search process, 
% For each triple in the training samples set, GMH finds the tail entity based on the head entity and relation through traversing in the background KG.
for each search step, the agent takes three operations: local-global knowledge fusion~(lines 5-6), differentiated action dropout~(line 7) and adaptive stopping search~(lines 8-10).
After finding the tail entity, the reward is calculated and the parameters are updated~(line 13).
Finally, the optimized parameters are output.

%Our experiments are discussed in the next section.

% \section{Experiment Setup}
\section{Experiment}
\label{sec:setup}

% \subsection{Dataset}
\subsection{Setup}
\label{sec:dataset}

\textbf{Dataset}
Existing popular benchmarks, such as UMLS~\cite{umls2007} and WN18RR~\cite{Dettmers2018Convolutional}, 
% mostly 
focus on the multi-hop reasoning in short\footnote{
    Considering that the current multi-hop reasoning works are concentrated on two or three-hop paths, we regard the path hop number greater than three as a long distance scenario.}
% \footnote{Considering the current state of multi-hop reasoning research area, we regard the distances smaller or equal to 3 as ``short'' ones and the distances larger than 3 as ``long'' ones.} 
distance scenarios.
% because their entity-pair distances are no more than three. 
Thus, they are unsuitable for evaluating %performance in 
complex cases requiring both long and short distance learning.
%of KGs. And no entity pair's distance is more than 3 in these common datasets.These datasets all focus on simple scenarios but lose sight of the complex scenarios that should exist.
To this end, we adopt the large-scale dataset FC17~\cite{neelakantan2015compositional} %which 
which contains
%two hundred thousand 
triples based on Freebase~\cite{freebase2008} enriched with the information fetched from ClueWeb~\cite{clueweb2013}.
% There are 46 query relation types in FC17.
%, however, the distance distribution in each task is extremely uneven. 
%In order to better promote the study of the long distance multi-hop reasoning task, 
%We introduce FC17-8, a subset of FC17 where the uneven tasks are removed.
%\footnote{The details on the dataset construction can be found at: \url{https://github.com/LongDistanceMultiHop/LongDist}}.
%There are 8 tasks in FC17-8's training/validation/testing sets. 
%These 8 relations are selected in the consideration of maximizing the number of long distance data.
%The triples in the training/validation/testing sets are classified by their distance and are excluded from the original KG. 
%Then, we intercept the data with distances are between 4 and 7 to generate the long version of FC8, which is denoted as FC8\_long. 
%The straightforward type of distance 1 and the infrequent type of distance larger than 8 are removed here.
%Note that the straightforward type of distance 1 is removed here.
Because the data with distance type larger than five is relatively small,
%and the amount of the required computing resources will increase dramatically as the distance increases,
we maintain the data with distance type between 2 and 5.
The sample number of each distance type~(2-5) %are 
is 63k, 53k, 11k, 5k, respectively.
Note that, there are extra relations served in the background KG plus 46 relation types in the train/valid/test sets of FC17.
We also evaluate our model on the other short distance datasets, i.e., UMLS and WN18RR. 
Table~\ref{table:datasets} summarizes the basic statistics of datasets.

% \begin{table}[t]
% % \floatbox[{\capbeside\thisfloatsetup{capbesideposition={left,top},capbesidewidth=4cm}}]{figure}[\FBwidth]
%         \caption{Statistics of datasets w.r.t. the number of entities and edges (the middle two columns) and the separation of the train/valid/test sets~(the right three columns).}
% \label{table:datasets} 
% \begin{center}
%         \resizebox{0.40\textwidth}{!}{
%                 % \resizebox{0.40\textwidth}{1.5cm}{
%                 \begin{tabular}{l|cc|ccc}
                
%                 \toprule[1.2pt]
%                 &   
%                 % $\mathcal{\left|E\right|}$  
%                 entity&     
%                 % $\mathcal{\left|R\right|}$    
%                 edge&    train   &     valid   &     test\\
%                 \midrule[0.6pt]
%                 FC17   &     49k    &   6k  &     125k   &     4k   &   5k\\
%                 UMLS   &     135    &   46  &     5k   &     652   &   661\\
%                 WN18RR   &     41k    &   11  &     87k   &     3k   &   3k\\
%                 % {\color{blue}NELL-995   &     75k    &   200  &     150k   &     543   &   4k\\}
%                 \bottomrule[1.2pt]
%                 \end{tabular}
%                 }
% \end{center}

% \end{table}

\begin{table}[t] 

    \centering 
    \caption{Statistics of datasets w.r.t. the number of entities and edges (the middle two columns) and the separation of the train/valid/test sets~(the right three columns).}
    %\vspace{-2mm}
\label{table:datasets}
   \resizebox{0.4\textwidth}{!}{
                % \resizebox{0.40\textwidth}{1.5cm}{
                \begin{tabular}{l|cc|ccc}
                
                \toprule[1.2pt]
                &   
                % $\mathcal{\left|E\right|}$  
                entity&     
                % $\mathcal{\left|R\right|}$    
                relation&    train   &     valid   &     test\\
                \midrule[0.6pt]
                FC17   &     49k    &   6k  &     125k   &     4k   &   5k\\
                UMLS   &     135    &   46  &     5k   &     652   &   661\\
                WN18RR   &     41k    &   11  &     87k   &     3k   &   3k\\
                % {\color{blue}NELL-995   &     75k    &   200  &     150k   &     543   &   4k\\}
                \bottomrule[1.2pt]
                \end{tabular}
                }
  %\vspace{0.3mm}
\end{table}

\begin{table*}[t]\centering   
  \footnotesize
  
  \caption{ MRR (\%) and HITS@N~(\%) scores~($\pm$ standard deviation) for multi-hop reasoning task on FC17, UMLS and WN18RR~(pairwise t-test at 5\% significance level). Higher values mean better performances and the best solution is marked in bold for each case.}
  %\vspace{-3mm}
  \centering
  \label{table:mainresults} 
  \resizebox{0.999\textwidth}{!}{
  \begin{tabular}{lcccc|p{0.5cm}p{0.5cm}p{0.5cm}|p{0.5cm}p{0.5cm}p{0.5cm}}
          \bottomrule[1.2pt]
  % \multirow{2}{*}{Model} & \multicolumn{4}{c|}{FC17}& \multicolumn{3}{c|}{UMLS}& \multicolumn{3}{c}{WN18RR} \\ 
  % & MRR& MRR($\leq 3$)&MRR($\geq 4$) & @1 &MRR & @1& @10&MRR & @1& @10\\ 
  \multirow{3}{*}{} & \multicolumn{4}{c|}{FC17} & \multicolumn{3}{c|}{UMLS}  & \multicolumn{3}{c}{WN18RR} \\ \cline{2-11}
& \multirow{2}{*}{MRR} & \multirow{2}{*}{MRR($\leq 3$)} & \multirow{2}{*}{MRR($\geq 4$)} &  \multicolumn{1}{c|}{HITS@N} & \multirow{2}{*}{MRR} & \multicolumn{2}{c|}{HITS@N} & \multirow{2}{*}{MRR} &  \multicolumn{2}{c}{HITS@N}
\\ \cline{5-5}\cline{7-8}\cline{10-11}
& & & &  @1  & & @1  &   @10  & & @1  &   @10 \\
  \hline\hline
          \rule{0pt}{12pt}
          TransE&11.91 {\footnotesize$\pm$0.20}&12.38 {\footnotesize$\pm$0.16}&9.37 {\footnotesize$\pm$0.28}&6.90 {\footnotesize$\pm$0.13 }&86.3&85.9&88.2&40.2&39.9&43.2\\\rule{0pt}{12pt}
Distmult&12.99 {\footnotesize$\pm$0.67 }&13.57 {\footnotesize$\pm$0.42 }&10.63 {\footnotesize$\pm$0.43}&8.53 {\footnotesize$\pm$0.71 }&86.8&82.1&96.7&46.2&43.1&52.4\\\rule{0pt}{12pt}
ComplEx&14.73 {\footnotesize$\pm$0.32 }&15.53 {\footnotesize$\pm$0.37}&10.91 {\footnotesize$\pm$0.54 }&9.68 {\footnotesize$\pm$0.53 }&93.4&89.0&99.2&43.7&41.8&48.0\\\rule{0pt}{12pt}
ConvE&18.98 {\footnotesize$\pm$0.63 }&19.75 {\footnotesize$\pm$0.63}&15.31 {\footnotesize$\pm$0.68 }&10.43 {\footnotesize$\pm$0.85 }&95.7&93.2&99.4&44.9&40.3&54.0\\\hline\rule{0pt}{12pt}
MINERVA&18.70 {\footnotesize$\pm$0.42 }&19.92{\footnotesize$\pm$0.36}&13.84{\footnotesize$\pm$0.45 }&9.08 {\footnotesize$\pm$0.94 }&82.5&72.8&96.8&46.3&41.3&51.3\\\rule{0pt}{12pt}
MultiHop&20.28 {\footnotesize$\pm$0.71}&21.63 {\footnotesize$\pm$0.65}&14.07{\footnotesize$\pm$0.76 }&10.53 {\footnotesize$\pm$0.84}&94.0&90.2&99.2&\textbf{47.2}&43.7&54.2\\\hline\rule{0pt}{12pt}
GMH&\textbf{23.75}{\footnotesize$\pm$0.52}&\textbf{25.06}{\footnotesize$\pm$0.52}&\textbf{18.52}{\footnotesize$\pm$0.56}&\textbf{12.98}{\footnotesize$\pm$0.76}&\textbf{96.2}&\textbf{93.9}&\textbf{99.9}&46.5&\textbf{45.3}&\textbf{55.8}\\

          \toprule[1.2pt]
  \end{tabular}}
%   \vspace{-4mm}
  \end{table*}

  \begin{table}[t]
    \centering
    \makeatletter\def\@captype{table}\makeatother
    \caption{MRR (\%) scores~($\pm$ standard deviation) for long distance multi-hop reasoning task~(pairwise t-test at 5\% significance level). The testing samples are divided into four types according to the distance.}
    %\vspace{-3mm}
    \label{table:longresults} 
    \resizebox{0.48\textwidth}{!}{
    \begin{tabular}{lcccc}
    \bottomrule[1.2pt]
    \multirow{2}{*}{} & \multicolumn{4}{c}{Distance Type} \\ \cline{2-5}
    & 4 & 5  & 6 & 7 \\ \hline\hline
    \rule{0pt}{12pt}
    ConvE&16.43{\footnotesize$\pm$0.59}&10.45{\footnotesize$\pm$0.53}&13.64{\footnotesize$\pm$0.63}&9.38{\footnotesize$\pm$0.97}\\\rule{0pt}{12pt}
    MINERVA&17.60{\footnotesize$\pm$0.73}&10.90{\footnotesize$\pm$0.61}&10.65{\footnotesize$\pm$0.88}&5.09{\footnotesize$\pm$0.87}\\\rule{0pt}{12pt}
    MultiHop&17.61{\footnotesize$\pm$0.62}&12.58{\footnotesize$\pm$0.89}&12.99{\footnotesize$\pm$0.85}&5.68{\footnotesize$\pm$0.95}\\\hline\rule{0pt}{12pt}
    GMH&{20.53}{\footnotesize$\pm$0.56}&{14.62}{\footnotesize$\pm$0.85}&{14.12}{\footnotesize$\pm$0.65}&{9.74}{\footnotesize$\pm$0.83}\\
    
    \toprule[1.2pt]
    \end{tabular}}
    %\vspace{-2mm}
    \end{table}

\noindent\textbf{Baselines}
% \subsection{Baselines}
\label{sec:baselines}

% We compare GMH %with the following approaches
% mainly with the family of the embedding-based models and the family of the multi-hop reasoning models as follows:\par
% \noindent{\tiny{$\bullet$}} \textbf{Embedding-based models:}
% TransE~\cite{bordes2013translating}, DistMult~\cite{yang2014embedding}, ComplEx~\cite{trouillon2017complex}, and ConvE~\cite{Dettmers2018Convolutional}.
% \noindent{\tiny{$\bullet$}} \textbf{Multi-hop reasoning models:} MINERVA~\cite{das2017go} and MultiHop~\cite{LinEMNLP2018}.
% The detailed descriptions of the above models can be found in Section~\ref{sec:related work}.
We compare GMH %with the following approaches
 with
%  the family of the embedding-based models and the family of the multi-hop reasoning models as follows: 
 1) the \textit{embedding-based models} involving
TransE~\cite{bordes2013translating}, DistMult~\cite{yang2014embedding}, ComplEx~\cite{trouillon2017complex}, and ConvE~\cite{Dettmers2018Convolutional};
as well as 
2) the \textit{multi-hop reasoning models} involving MINERVA~\cite{das2017go} and MultiHop~\cite{LinEMNLP2018}.
% The detailed descriptions of the above models can be found in Section~\ref{sec:related work}.

% \subsection{Implementation Details}
\noindent\textbf{Implementation Details}
\label{sec:implementation}
GMH is implemented on PyTorch and runs
% , trained and tested 
on a single TITAN XP. Following~\cite{das2017go}, we augment KG with the reversed link $(e_{t},r^{-1},e_{h})$ for each triple.
We exclude the triples from the training set if they occur in the validation or testing sets.
% for the sake of the overlap between the sets.
For the baselines and GMH, we set the maximum search step~$S$ to five because the entity pair's distance is up to five in FC17.
For the short distance datasets, UMLS and WN18RR, $S$ is set to three.
The maximum loop number~$N$ for all datasets is set to two.
%The activation functions are
We employ softmax function as the activation function.
All hyper-parameters are tuned on the validation set can be found in supplementary materials 
% We set the relation and entity embedding dimension size to 200, the learning rate equal to 0.003 and the batch size to~(128-512).
% The settings of other parameters are consistent with those in \cite{LinEMNLP2018}, while the hyperparameters are tuned on the validation set.
The pre-trained embedding-based model that we adopt is ConvE.
We optimize all models with Adam~\cite{adam}\footnote{We will release the processed dataset and source code, after the paper is published.
The description of datasets and other details can be found in supplementary materials. 
}.

%We fix the batch size to 128, and set the maximal epochs to 500 for the embedding-based models and 150 for the multi-hop reasoning models and our model. 

\noindent\textbf{Metrics}
We follow the evaluation protocol of~\cite{LinEMNLP2018} %which 
that records the rank of the available entities at final step in a decreasing order of confidence score for each query, and adopts mean reciprocal rank~(MRR) and HITS@N to evaluate the results.
All results given %by our 
in our experiments %were 
are the mean and standard deviation values of three training repetitions.

\subsection{Multi-hop Reasoning}
\label{sec:results}

Table~\ref{table:mainresults} shows the results obtained on FC17 and two short distance datasets, UMLS and WN18RR based on MRR (\%) and HITS@N (\%) measures.
% To statistically measure the significance of the performance difference, the pairwise t-test at 5\% significance level is conducted between the algorithms.
On the FC17 dataset, GMH achieves 23.75\% MRR score surpassing the second-best model MultiHop with 3.47\% improvement based on the MRR metric. This includes 3.43\% improvement on short distance samples and 4.45\% improvement on long distance samples.
%The performance of 
We observe that multi-hop reasoning models outperform most embedding-based models, but their performance declines when the distance increases. We assume this may be attributed to the significantly increasing difficulty of building long paths when predicting long distance relations. 
The embedding-based models appear to be less sensitive to the distance variations, but they neglect the deep information existing in multi-hop paths, which limits the interpretative ability of predicting results. 
We further evaluate the short-distance reasoning performance on UMLS and WN18RR.
The results of %UMLS and WN18RR datasets
the baselines
are cited from~\cite{LinEMNLP2018}. 
% As shown in Table~\ref{table:mainresults}, GMH performs comparably well in reasoning the short-distance relations, while is not as predominant as its %exhibitions 
% performance in the long-short compound reasoning cases or long distance reasoning cases. 
GMH performs comparably well in reasoning in the short distance scenarios, yet its effectiveness in the long-short compound reasoning or long distance reasoning scenarios is more obvious.
%Thus, %We assume this may because the short distance reasoning case is too simple for GMH to really work, while 
For the WN18RR dataset, GMH performs weaker than MultiHop.
We speculate that this is because the number of relations in WN18RR is extremely smaller than the number of entities, which will make it difficult to accurately learn the relation embeddings.
Choosing a superior pre-trained embedding-based model is critical for our model.

% We argue that GMH is more effective in the long-short compound distance cases which are typical in real-world application scenarios.

% \input{body/align_table}

\noindent
\textbf{Multi-Hop Reasoning in long distance scenarios}
%To exploring GMH's performance on the longer distance~(6 and 7) situation,
%we evaluate it on the small dataset, FC17\_8, a subset of FC17.
As we noticed in Table~\ref{table:mainresults}, GMH % 
achieves new state-of-the-art results 
%shows promise 
on FC17 %which
dataset which contains both short distance and long distance types.
We further evaluate its performance in terms of reasoning on the relations in 
longer distances,
%Because our work focuses on the long distance multi-hop reasoning task, 
which have been rarely examined by the existing works.
%we also evaluate our model on longer distance instances~(i.e., 4-7) on FC17-8, a subset of FC17,
Therefore, we extract the relations from FC17 whose distances span from 4 to 7 and in this way we construct a subdataset, called FC17-8, which
contains eight query relation types
% \footnote{The eight relation types in FC17-8 are: (1) broadcast-content-genre, (2) business-industry-companies, (3) cvg-game-version-game, (4) cvg-game-version-platform, (5) music-artist-origin, (6) organization-organization-locations, (7) people-deceased-person-cause-of-death, (8) tv-tv-program-country-of-origin.}
. %and the most longest distance samples
%There are 46 tasks (i.e., the type of query relations) in FC17, while the long distance~(i.e., $>$ 5) distributions are extremely uneven in some tasks.
Table~\ref{table:longresults} displays the results of reasoning on the four distance types based on the MRR metric.
Compared with GMH and the multi-hop reasoning models, the embedding-based model seems less sensitive to the distance variations, %. Yet 
while its reasoning performance is inferior to %our model and the multi-hop reasoning models 
the compared models on all distance types. 
GMH consistently yields the best performance on the long distance reasoning scenarios.
%The top-3 improvement cases are respectively the distance 4, distance 5, and distance 7, which indicate the superiority of our model for the long distance link prediction task.
We observe that all the models perform better on the even distance type~(4 and 6) than odd distance type~(5 and 7).
There are two possible reasons:
1) there is an imbalance between the difficulty and the number of different distance types;
2) the models are better at reasoning on symmetric paths like the four-hop path~
\textit{Stephen~Curry}
$\xrightarrow[]{plays~for}$
\textit{Golden~State~Warriors}
$\xrightarrow[]{compete~in}$
\textit{NBA}
$\xleftarrow[]{compete~in}$
\textit{Houston~Rockets}
$\xleftarrow[]{plays~for}$
\textit{James Harden}.

In addition to the %striking 
superior reasoning capability of GMH as demonstrated in Table~\ref{table:mainresults} and Table~\ref{table:longresults}, other promising potentials pave the way for GMH in advanced applications. %On the one hand
First, GMH is explainable because it considers the path information, which is beyond the scope of the existing embedding-based models. %On the other hand
Second, the global knowledge learnt from graph structure, which has been overlooked by the existing multi-hop reasoning models, is incorporated in GMH.

% \subsection{Discussion}
% \label{sec:discussion}

\subsection{Analysis of GMH}
\label{sec:ablation}

\begin{figure}
  
  \centering
  \includegraphics[height=1.3in]{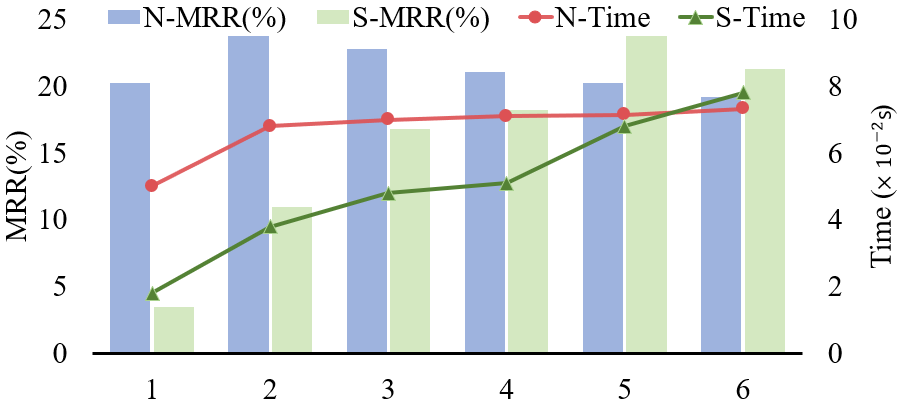}
  \caption{MRR~(\%) scores and running time of GMH over different (a) maximum search step~$S$ and (b) maximum loop number~$N$ on FC17. Best viewed in color.}
  \label{fig:bandwidth}
% \vspace{2mm}
\end{figure}

\begin{figure}[t]
    \centering 
    \subfigure[Scoring Function Type]{
    \begin{minipage}[t]{0.46\linewidth}
    \centering
    \centerline{\includegraphics[height=1in]{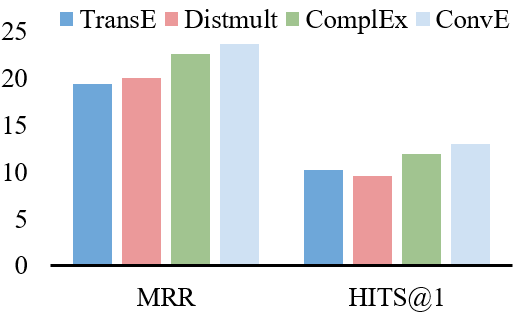}}
    \end{minipage}
    \label{fig:abalation1}
    }%
    \subfigure[Aggregator Type]{
    \begin{minipage}[t]{0.49\linewidth}
    \centering
    \centerline{\includegraphics[height=1in]{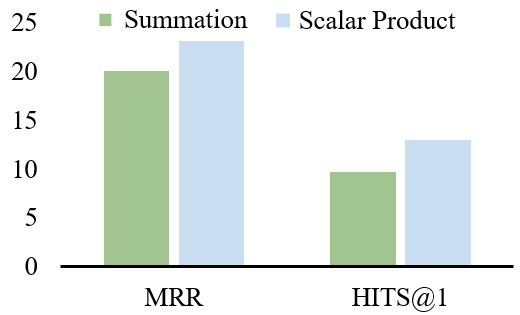}}
    \end{minipage}
    \label{fig:abalation2}
    }%
    \centering
    %\vspace{-3mm}
    \caption{MRR (\%) and HITS@N (\%) scores comparison of GMH over different scoring functions and aggregators on FC17. Best viewed in color.}
    % \vspace{2mm}
    \label{fig:abalation_main}

\end{figure}

In this section, we conducted an extensive analysis of GMH from two aspects:
1) modules~(c.f., Table~\ref{table:ablation});
2) hyper-parameters~(c.f., Figure~\ref{fig:bandwidth});
and 3) scoring functions and aggregators (c.f., Figure~\ref{fig:abalation_main}).
% Firstly, we evaluate the contributions of individual components of GMH model, as shown in Table~\ref{table:ablation} and Figure~\ref{fig:abalation_main}.

\noindent
\textbf{Local Knowledge vs. Global Knowledge}
We fuse two components (i.e., the local knowledge $lk_{s}$ and the global knowledge $gk_{s}$) to enable the search agent to find the target tail entity. Thus, an extensive experiment is conducted to test the contributions of $lk_{s}$ and $gk_{s}$ in the multi-hop reasoning task.
% Three strategies are considered here:
% (a) using only the local knowledge (``onlyL''), (b) using only the global knowledge (``onlyG''), and (c) combining two factors (``L+G").
The top three lines of Table~\ref{table:ablation} reveal that 
% As shown in Table~\ref{table:ablation}, 
fusing $lk_{s}$ and $gk_{s}$ achieves the best results under different evaluation metrics. 
Removing %any 
either knowledge %will result in 
yields a significant performance drop. 
Concretely, removing the local knowledge causes a 9.10\% MRR degradation, %while 
and removing the global knowledge results in a 4.05\% MRR degradation.
%Thus
This suggests that the local knowledge %is
may be more beneficial for the search agent than the global knowledge, and using only the %global embedding 
local knowledge to find a path in KG %is 
may be ineffective in the training process.
%However, 
Still we argue that the importance of the global knowledge should not be neglected, especially when it is combined with the local knowledge to handle the ``where to go'' issue.

\begin{table}

  \caption{Analysis of GMH on FC17: GKL, LKL, DAD, and ASS represent the global knowledge learning, local knowledge learning, differentiated action dropout, and adaptive stopping search respectively.}
  \label{table:ablation}
  %\vspace{-2mm}
  \centering\scriptsize
  \resizebox{0.49\textwidth}{!}{
  \begin{tabular}{lccc}
  \bottomrule[0.8pt]
  \multirow{2}{*}{} & \multirow{2}{*}{MRR} &  \multicolumn{2}{c}{HITS@N} \\ \cline{3-4}
  & & @1  & @10 \\ \hline\hline
  GKL&13.28&6.47&13.39\\
  %\rule{0pt}{12pt}
  LKL&18.33&9.86&25.17\\
  %\rule{0pt}{12pt}
  LKL+GKL&22.38&11.65&28.05\\
  %\rule{0pt}{12pt}
  % LPF+GEF+RAD&22.84&12.39&28.81\\\hline
  LKL+GKL+DAD&23.25&12.10&28.29\\\hline
  % LPF+GEF+DPD+ASS(GMH)
  GMH (LKL+GKL+DAD+ASS)&23.75&12.98&29.86\\
  \toprule[0.8pt]
  \end{tabular}}

\end{table}

\noindent
\textbf{Performance~w.r.t.~Differentiated Action Dropout}
The differentiated action dropout module is adopted to increase the diversity of search paths in the training stage.
The fourth line of Table~\ref{table:ablation} shows the validity of this module.
We also test the effect of randomly action dropout~(22.15\% under MRR), and there is a gap with our model.
This illustrates that the reason why the differentiated action dropout 
performs well is because the mask operation is based on the global knowledge rather than on random strategy.

\noindent
\textbf{Performance w.r.t.~Adaptive Stopping Search}
As mentioned before, we have devised the adaptive stopping search module to avoid 
wasting of resources caused by over-searching, i.e., the ``when to stop'' issue.
As can be seen from the bottom two rows of Table~\ref{table:ablation}, ASS also has a slight effect on the performance.
% This is because that this module can partly prevent the search agent from missing the target tail entity.
This is because the module can partially prevent the search agent from continuing to search when the target tail entity has been found.

% %\vspace{1mm}
% In the last experimental analysis, we evaluate the hyper-parameters of GMH~(i.e., maximum search step~$S$ and maximum loop number~$N$) on MRR~(\%) and running time~(second/sample). 

\noindent
\textbf{Maximum Search Step}
As shown in Figure~\ref{fig:bandwidth}, GMH achieves best performance at~$S=5$. %Obviously
Using a large~$S$ will cause wasting of resources, while if using a small~$S$, it will affect the performance on the long distance reasoning samples.
Meanwhile, the running time rises sharply when increasing~$S$.
Therefore, the introduction of adaptive stopping search module is necessary and rational.

\noindent
\textbf{Maximum Loop Number}
We divide the self-loop action into two types: positive and negative. The positive self-loop action means the agent arrives at the target tail entity, while the negative self-loop action means the current entity is not the target.
See Figure~\ref{fig:bandwidth}, 
%the little~$N_{ESS}$ will misjudge negative actions as positive actions and the big~$N_{ESS}$ will do the opposite and lose the effectiveness of ESS 
a small~$N$ may cause the agent to misrecognize negative actions as positive actions, while a large~$N$ may lead to 
% the opposite effects but lose the effectiveness of adaptive stopping search module.
lose the advantage of reducing time consumption.
Compared with not using the adaptive stopping search module~(i.e.,~$N=1$), using it has resulted in a significant improvement with the optimal number of 2.

% Secondly, we conducted experiments to examine the effectiveness of different scoring functions and aggregators, as shown in Figure~\ref{fig:abalation_main}.

\noindent
\textbf{Scoring Function Types}
The pre-trained embedding-based model that we adopt is ConvE. For more extensive ablation analysis, we have conducted the experiments by incorporating effective embedding-based models~(i.e., TransE, DistMult, ComplEx, and ConvE).
As shown in Figure~\ref{fig:abalation1}, 
ConvE has a superb ability to learn the global semantic representation than other embedding-based models.
% %the performance of 
% using %these 
% other embedding models makes subtle difference than %that of 
% using ConvE. %which performs best. 
% We adopt ConvE as the pre-trained model since ConvE has a superb ability to learn the global semantic representation.

\noindent
\textbf{Aggregator Types}
We next investigate the performance of our model
w.r.t different aggregator types. We adopt two types of aggregators: summation and scalar product, to fuse the local knowledge~$lk_{s}$ and global knowledge~$gk_{s}$.
We can see from Figure~\ref{fig:abalation2} that the scalar product outperforms the summation. The advantage of the scalar product aggregator is that the multiplication operation can increase the discrimination between available actions.

% \vspace{-2mm}
\section{Conclusions}
\label{sec:conclusions}

% In this study, we %have focused 
% focus on the long distance-aware multi-hop reasoning task that has rarely been %ignored 
% approached in the previous studies. %We have proposed a 
% Our proposed %novel 
% model %, called 
% GMH, which seamlessly unifies both local path factors and global embedding factors, %, to approach this task which yields considerable performance improvements on both short and long distance scenarios.
% can successfully accomplish this task as demonstrated by extensive experiments.
% %the newly constructed dataset.
% %Our future work will consider incorporating external text data and developing our model to solve further complex tasks.
We have studied the multi-hop reasoning task in long distance scenarios and proposed a general model which could tackle both short and long distance reasoning scenarios.
Extensive experiments showed the effectiveness of our model on three benchmarks.
% We hope insights from this work will inspire future developments of KG completion and KG-related applications.
We will further consider the feasibility of applying our model to complex real-world datasets with more long distance reasoning scenarios and more relation types.
% The superior performance of our model in long distance reasoning scenarios provides opportunities for conducting complex reasoning on KG-related applications, e.g., complex question answering based on KG.
Besides, we have noticed that there are other ``interference'' in long distance reasoning. For example, noise from the KG itself, i.e., the fact that it lacks validity. These noises can gradually accumulate during long distance reasoning and affect the result confidence. We leave the further investigation to future work.

\section{Acknowledgements} 
    We sincerely thank Jun Wang and Xu Zhang for their constructive suggestions on this paper.
    This work was supported by the China Postdoctoral Science Foundation (No.2021TQ0222).

\bibliography{anthology,custom,ref}
\bibliographystyle{acl_natbib}

\appendix

% \section{Example Appendix}
% \label{sec:appendix}

% This is an appendix.

\end{document}